\documentclass{article}

\usepackage[preprint]{corl_2025} 

\usepackage{amsmath,amsfonts,bm}

\def\Figref#1{Figure~\ref{#1}}

\def\Secref#1{Section~\ref{#1}}

\def\eqref#1{equation~\ref{#1}}

\def\1{\bm{1}}

\def\va{{\bm{a}}}

\def\vg{{\bm{g}}}
\def\vh{{\bm{h}}}

\def\vo{{\bm{o}}}
\def\vp{{\bm{p}}}
\def\vq{{\bm{q}}}

\def\vs{{\bm{s}}}

\def\vx{{\bm{x}}}

\def\mB{{\bm{B}}}

\def\mR{{\bm{R}}}

\DeclareMathAlphabet{\mathsfit}{\encodingdefault}{\sfdefault}{m}{sl}
\SetMathAlphabet{\mathsfit}{bold}{\encodingdefault}{\sfdefault}{bx}{n}

\def\sR{{\mathbb{R}}}

\input{acronyms}

\usepackage{graphicx}
\usepackage{cleveref}
\usepackage{amssymb}
\usepackage{booktabs}
\usepackage{subcaption}

\usepackage{wrapfig}

\newtheorem{assumption}{Assumption}

\title{
Towards Safe Robot Foundation Models\\
Using Inductive Biases}

\author{
  Maximilian T\"olle$^{\star, 1, 2}$\; Theo Gruner$^{\star, 1, 3}$\; Daniel Palenicek$^{\star, 1, 3}$ \\
  \textbf{Tim Schneider}$^{1}$:  \textbf{Jonas G\"unster}$^{1}$\; 
  \textbf{Joe Watson$^{4}$\; Davide Tateo$^{1}$\; Puze Liu$^{1, 2}$\; Jan Peters$^{1, 2, 3, 5, 6}$}\\
  $^{\star}$Equal contribution\; $^{1}$Technical University of Darmstadt\; $^{2}$German Research Center for AI (\textsc{dfki})\\
  $^{3}$hessian.AI\; $^{4}$University of Oxford\; $^{5}$Robotics Institute Germany (\textsc{rig})\; $^{6}$Centre for Cognitive Science
}

\begin{document}
\maketitle

\begin{abstract}
    Safety is a critical requirement for the real-world deployment of robotic systems.
    Unfortunately, while current robot foundation models show promising generalization capabilities across a wide variety of tasks, they fail to address safety, an important aspect for ensuring long-term operation.
    Current robot foundation models assume that safe behavior should emerge by learning from a sufficiently large dataset of demonstrations.
    However, this approach has two clear major drawbacks.
    Firstly, there are no formal safety guarantees for a behavior cloning policy trained using supervised learning.
    Secondly, without explicit knowledge of any safety constraints, the policy may require an unreasonable number of additional demonstrations to even approximate the desired constrained behavior. 
    To solve these key issues, we show how we can instead combine robot foundation models with geometric inductive biases using \acs{atacom}, a safety layer placed after the foundation policy that ensures safe state transitions by enforcing action constraints.
    With this approach, we can ensure formal safety guarantees for generalist policies without providing extensive demonstrations of safe behavior, and without requiring any specific fine-tuning for safety.
    Our experiments show that our approach can be beneficial both for classical manipulation tasks, where we avoid unwanted collisions with irrelevant objects, and for dynamic tasks, such as the robot air hockey environment, where we can generate fast trajectories respecting complex tasks and joint space constraints.
    For experimental results, see
    \url{https://sites.google.com/view/safe-robot-foundation-models}.
\end{abstract}

\keywords{robot foundation models, generalist policies, safety} 

\section{Introduction}
\Acp{rfm}~\cite{open_x_embodiment_rt_x_2023, kimOpenVLAOpenSourceVisionLanguageAction2024, octo_2024, pi_0} have shown a promising direction to carry out a large set of tasks across a wide range of robotics systems using textual commands as task descriptions.
The current developments of \acp{rfm} focus on enhancing the generalization across tasks, data modalities, and robot embodiments.
In contrast, other key practical properties of robotic systems, such as safety, are not considered in depth.
The key assumption behind this choice is that the emergent behavior will be inherently safe since \acp{rfm} are trained using \ac{bc}.
If the data distribution only contains safe trajectories, one would expect the learned behavior to also be safe.
However, while this assumption may hold in simple pick-and-place scenarios, we argue that in general settings and dynamic environments, this assumption may be limiting for many key reasons. Firstly, there is no way to ensure any formal theoretical guarantees that a \ac{rfm} will satisfy any safety constraint.
Scaling the amount of data will move us closer and closer to the demonstrator distribution, but this does not guarantee that the policy will not generate dangerous behavior.
Secondly, often safety constraints such as collision avoidance, joint limits, or any other geometric constraints are both common in robotics applications and can be easily enforced. While it would be possible to learn a policy able to deal with these safety constraints, it may require an unreasonable amount of data, particularly to make the policy robust to out of distribution objects, distractors, or unexpected disturbances of the workspace, such as a human or an animal entering the working area of the robot.
In some cases, generating this data would not be even possible, due to the danger of allowing living beings in the work area of the robot, particularly if the robot's embodiment is not intrinsically safe, as happens in the setting of self-driving cars or heavy industrial robots.

\begin{figure}[t]
    \centering
    \includegraphics[width=\linewidth]{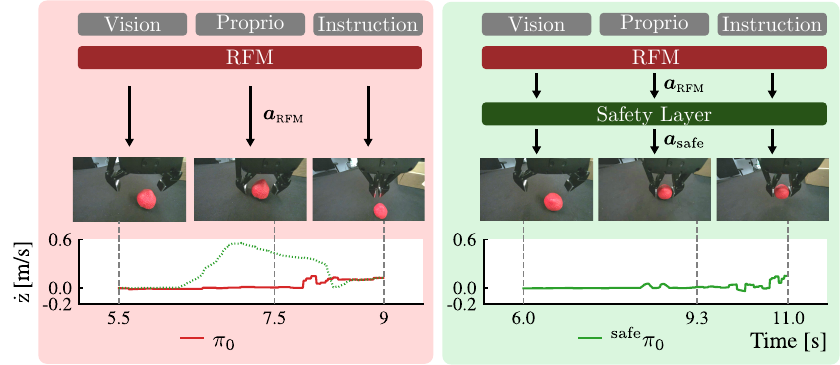}
    \caption{Our proposed safety layer can be added to the output of an arbitrary \ac{rfm}, e.g., $\pi_0$. \\(left) Without the added \ac{atacom} safety layer, the vanilla \piZero{} policy crashes the robot into the table. We highlight the importance of the safety layer by plotting the impact of the \piZero{} action on the end effector's vertical velocity. While vanilla \piZero{} corrects the z position after the impact (red), the safety layer would have engaged earlier to circumvent crashing into the table. (right) Deployment with the safety layer results in a safe rollout without pronounced z-correction.}
    \label{fig:unsafe_behavior}
\end{figure}

For this reason, we believe that empowering robots with the knowledge of safety constraints and enforcing them directly at the policy level is extremely beneficial both in terms of formal safety guarantees --- fundamental for the acceptance of robotics systems in society ---and in terms of amount of data and computation needed to operate safely.
To reach this objective, we rely on the key concept of inductive biases: instead of collecting demonstrations to account for the full space of possible solutions that \acp{rfm} can generate, we restrict our policy to only generate safe trajectories.
Furthermore, we can rely on the fact that robots interact with the physical world, where the geometry of the rigid objects plays a crucial role.
Using the information coming from the object geometry, which can be easily extracted from off-the-shelf perception pipelines, we can robustly generate safety geometric constraints that can be combined with existing prior knowledge about the robot to ensure safety with formal guarantees.

To achieve this ambitious goal, we propose a modular safety layer based on the \ac{atacom} approach~\cite{liu2022_atacom, liu2024_atacom_journal}, which translates known safety constraints into a constraint manifold to ensure safe actions.
Due to the design of \ac{atacom}, we can easily combine this safety layer with any \ac{rfm} and enforce safe behavior w.r.t. a given set of safety constraints.
Furthermore, assuming proper constraint evaluation \ac{atacom} provides safety guarantees~\cite{liu2024_atacom_journal} in terms of forward invariance and input-to-state stability w.r.t. the safe set.

To further demonstrate the generality of this approach, we present a semi-automated pipeline based on \ac{sam2}~\cite{ravi_2024_sam2} to automatically build box constraints for collision avoidance with any object. This allows our method to go beyond user-defined constraints, e.g., workspace constraints or self-collision avoidance. While our proposed solution is not a fully-fledged automatic constraint generation methodology, it shows that this module could be, in principle, implemented robustly and reliably.

We extensively evaluate our approach both in simulation and in real-world platforms, using two different \acp{rfm}, \piZero{} and \ac{octo}. Specifically, we focus on two different platforms, solving two very different types of tasks: a classical manipulation setup using the Franka robot, and a dynamic task, where we train a Kuka IIWA robot to perform a hitting motion on the air hockey game.
We want to prove that our methodology is both flexible and computationally efficient while ensuring safety.
Our results show that our safety layer approach, despite the lack of specific fine-tuning data, is not considerably affecting the success rate, while ensuring safe policy execution both in the dynamics and quasistatic tasks.


\section{A Safety Module for Robot Foundation Models} 
In the following, we explain in detail how to combine an arbitrary \ac{rfm} with the \ac{atacom} safety layer. Our approach follows a different framework w.r.t. existing approaches,
as most approaches neglect the safety issue, assuming that enough safe data is provided to the model.
Prior attempts at safety perform post-training alignment of \ac{rfm}s to make the models \textit{safer}~\cite{zhang2025safevla}. In contrast to these approaches, we seek to make the model inherently safe on an architectural level, under the assumption of known explicit safety constraints. In principle, it is possible to use an arbitrary safety layer, however, \ac{atacom} is a good choice in terms of robustness and simplicity.

To achieve safety through the \ac{atacom} safety layer, we require two key assumptions.
\begin{assumption}
    \label{asmp:affine_system}
    Access to the system's state $\vs$ and a control affine system $\dot\vs = f(\vs) + G(\vs)\va$.
\end{assumption}
This assumption appears quite restrictive as we require access to the model of the robot and its state.
However, for most tasks considered so far for \ac{rfm}s, only the kinematic model is required, and therefore, many settings of interest can be described in terms of control-affine systems.
Moreover, even if kinematic knowledge is not strictly required by the foundation model, this information is normally used further down the control stack and therefore readily available.

\begin{assumption}
    \label{asmp:constraints}
    The safety conditions can be described as continuously differentiable constraints $\bm{0} \geq g(\vx)\in\mathcal{C}^1$. The constraint function should be known analytically.
\end{assumption}
While this assumption is quite strong, it holds in most practical scenarios, particularly when safety is critical.
In general, some constraints are trivial to impose, e.g., workspace limits. 
We show in Section~\ref{sec:visual_constraints} how we can easily generate safety constraints for simple visual manipulation tasks. Similar approaches can be used to generate arbitrary constraints, or it could be possible to exploit common-sense knowledge coming from a general-purpose large language model. 
In general, to generate arbitrary safety constraints, it is necessary to learn them automatically from environment interaction.
This problem is an active research topic, and many solutions already exist in the literature~\cite{taylor2020learning,tan2023your,yang2023model}, including how to deal with constraint uncertainty~\cite{yang2021wcsac,guenster2024handling}.
However, in this paper, we consider only the setting where constraints are known, leaving more advanced automatic approaches for future work.

\subsection{Acting on the Tangent Space of the Constraint Manifold}
Under assumptions~\ref{asmp:affine_system} and~\ref{asmp:constraints}, we can couple any arbitrary \ac{rfm} with the \ac{atacom}~\citep{liu2022_atacom} safety layer.
The key idea of \ac{atacom} is to generate a safe action space where we can sample arbitrary actions, while ensuring the satisfaction of the safety constraints. 
This satisfaction is achieved by constructing the so-called constraint manifold, computing the tangent space at the current robot configuration, and using this tangent space as a safe action space.
By taking actions on the tangent space of the constraint manifold, we generate paths moving on the manifold, corresponding to safe trajectories of the robot.
Under mild assumptions, this approach ensures theoretical guarantees in terms of forward invariance of the constraint manifold and input-to-state stability~\cite{liu2024_atacom_journal}, ensuring that safety is achieved even under disturbances.

The \ac{atacom} safety layer takes as input an arbitrary action, i.e., the action $\va_{\mathrm{RFM}}$ sampled from the \ac{rfm}, and produces a safe action to apply to the system by constraining the input action when necessary.
We refer to \citet{liu2024_atacom_journal} for the technical details.
At a high level, the \ac{atacom} action can be decomposed into three components as follows
\begin{equation}
    \va_{\mathrm{safe}} = \va_{\mathrm{drift}}(\vs) + \va_{\mathrm{err}}(\vs) + \mB(\vs)\va_{\mathrm{RFM}},
\end{equation}
where $\va_{\mathrm{drift}}(\vs)$ is a state-dependant compensation term that compensates for the change of the constraint function due to the affine component of our system (the "drift" of the system), $\va_{\mathrm{err}}(\vs)$ is an additional error correction term that is active only in case of constraint violations (e.g., in case of disturbance), bringing the system back to the safe set.
The last component represents the tangent space basis $\mB(\vs)$ in the current state, having the effect of morphing the action sampled from the \ac{rfm} to avoid constraint violations.
While the \ac{atacom} action space can, in principle, have a different physical meaning w.r.t. the vanilla action space, the $\mB$ matrix can be chosen such that the morphing retains as much as possible the action semantics.
This allows us to combine the safety layer with any \ac{rfm} without performing a safety-layer specific fine tuning.

Another important issue is that satisfying safety constraints may require a higher control frequency than what is possible with current \ac{rfm} due to their slower latency compared to simpler robot policies.
This requirement is because the safety constraint depends on the robot's state, which may evolve at a faster timescale than the action frequency.
However, the computation needed for the \ac{atacom} safety layer can be performed at a much higher frequency, as we only require computing the drift term, the error correction term, and the basis matrix $\mB$ using the state and the analytical constraints.
In practice, we sample the action from the \ac{rfm} at a fixed rate, e.g., 15Hz, repeating the single action or predicting a series of actions.
The action is then applied to the \ac{atacom} layer, where the basis, error correction, and drift compensation terms change at a higher frequency, e.g., 60Hz.

Using the \ac{atacom} framework, we can impose a wide variety of constraints.
Several core safety constraints can be defined using only the robot's kinematic model to ensure a general notion of safety.
\emph{Joint limit constraints} prevent the robot from exceeding its mechanical bounds while still allowing control near those limits, enabling high flexibility.
\emph{Workspace constraints} allow the positioning of cameras and other essential components within the scene.
\emph{Self-collision constraints} are implemented by approximating safety-critical parts of the robot with geometric volumes (e.g., spheres) that enclose the mesh.
\ac{atacom} efficiently handles a large number of constraints through parallelization of the constraint computations. 
If the geometry of the obstacles in the environment is available, it is possible to impose complex \emph{collision avoidance constraints}.
This capability can be achieved by defining the constraint as a \ac{sdf}~\cite{liu2022regularized,liu2023safe}.
However, if it is not necessary to operate with extreme precision around a given obstacle, it is always possible to define a bounding box as constraints around the object.
In the next section, we will explain how to easily generate bounding box constraints from visual input.

\subsection{Visual Constraint Generation}
\label{sec:visual_constraints}

\begin{figure}[t]
    \centering
    \begin{subfigure}[t]{0.32\textwidth}
        \includegraphics[width=\linewidth]{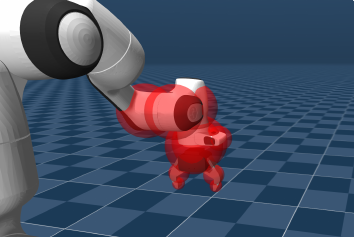}
    \end{subfigure}
    \begin{subfigure}[t]{0.32\textwidth}
        \includegraphics[width=\linewidth]{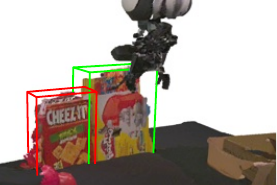}
    \end{subfigure}
    \begin{subfigure}[t]{0.32\textwidth}
        \includegraphics[width=\linewidth]{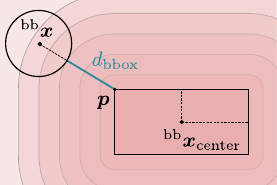}
    \end{subfigure}
    \caption{(left) Spheres cover the robot's hull at critical areas to formulate distance-based constraints ensuring safe executions of the \acp{vla} action predictions. (middle) Bounding boxes of obstacles are generated from 2D instance segmentation and depth information. (right) We calculate the distance between the covering spheres and the obstacle's bounding box by projecting the sphere's center into the bounding box's coordinate frame and estimating the distance to the bounding box's hull.}
    \label{fig:robot_spheres}
\end{figure}

Manual definition of constraints for the \ac{rfm} is time-consuming and often requires expert knowledge and environmental information.
State-based environments offer all the necessary information out of the box to specify safety constraints.
However, \ac{rfm}s usually only observe visual information about the environment they operate in order to generalize to arbitrary real-world environments.
As such, occlusions can lead to partial observability, and the 2D camera stream needs to be mapped into the 3D task space of the robot to infer and define constraints effectively.
While the definition of static workspace constraints and joint constraints is trivial, the definition of constraints for non-static objects in the scene is tedious and non-trivial.  
Technologies like OptiTrack could be used, but reliance on such specialized hardware would pose a significant limitation and only work in lab environments.
As such, we provide an intuitive, cost-effective, and lightweight approach to automatic constraint generation in the visual space.  

We leverage instance segmentation in 2D using \ac{sam2}~\cite{ravi_2024_sam2} and lift obtained multi-view segmentation masks into 3D using the pinhole camera equation and the camera intrinsics. Based on the 3D instance segmentation, we calculate minimum bounding boxes for each object in the scene. 
We draw the bounding boxes ourselves to obtain reliable bounding boxes for every evaluation run. This process can be automated in future work by incorporating more grounded segmentation masks, for example, following the approaches in~\cite{liu_grounding_dino_2024, ren_grounding_dino_1.5_2024}.

We calculate the distance between points on the mesh of the robot and the bounding box planes. 
We parameterize the oriented bounding box (bb) through its center in the world frame ${}^{\mathrm{base}}\vx_{\mathrm{center}}$, its rotation ${}^{\mathrm{base}}\mR_{\mathrm{bb}}$, and its extent in the bounding boxes frame ${}^{\mathrm{bb}}\vh$.
To obtain distance estimates between the robot and the segmented objects, we spawn spheres at key robot positions that cover the manipulator's hull (\Figref{fig:robot_spheres}).
Each sphere is parameterized by its center position and radius $(\vx, r)$. To ensure safety, \ac{atacom} guarantees that the distance between each sphere's hull and the obstacle's bounding box remains positive. This is done by projecting the sphere's center into the oriented bounding box frame ${}^{\mathrm{bb}}\vx = {}^{\mathrm{bb}}\mR_{\mathrm{base}} {}^{\mathrm{base}}\vx + {}^{\mathrm{bb}}\vx_{\mathrm{center}}$. 
We then calculate the distance between the sphere and the closest point on the bounding boxes surface
\begin{align}
    d_{\mathrm{bb}} = || \alpha{}^{\mathrm{bb}}\vx - \vp\|; \quad \vp_i = \mathrm{clip}( \alpha {}^{\mathrm{bb}}\vx_i, -{}^{\mathrm{bb}}\vh_i / 2, {}^{\mathrm{bb}}\vh_i / 2); \quad\alpha = 1 - r/ \| {}^{\mathrm{bb}}\vx\|.
    \label{eq:bbox_constraint}
\end{align}
Here, $\vp$ denotes the closest point of the bounding box to the sphere and can be obtained by clipping the sphere's reach onto the bounding box's limits, and $\alpha$ projects the sphere's center to its boundary. With that, we have a simple and effective method to estimate a bounding box constraint $g_{\mathrm{bb}}(\vx) = - d_{\mathrm{bb}}$ with an analytical gradient estimation.

\section{Experiments}

\begin{wrapfigure}[17]{r}{0.45\textwidth}
  \vspace{-1.5em}
  \begin{center}
    \includegraphics[width=0.45\textwidth]{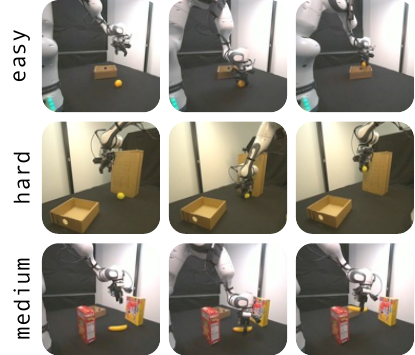}
  \end{center}
  \caption{Video frame extracts from a rollout on three different tasks.
  Difficulty is dictated by the number of obstacles in the scene. }
  \label{fig:franka-setup}
  \vspace{-3em}
\end{wrapfigure}

We provide several experiments in two different environments to evaluate the behavior of \acp{rfm} from a safety perspective. First, we set up a quasi-static pick-and-place environment, most commonly represented in current large-scale training datasets~\cite{open_x_embodiment_rt_x_2023, droid_2024}. Here, a Franka Research 3 (\textsc{fr3}) needs to be controlled to accomplish the task of picking and placing various items while not colliding with the workspace or obstacles within the workspace~(\Cref{fig:franka-setup}). Next, we use the air hockey environment~\cite{air_hockey_challenge} to showcase the highly dynamic task of puck hitting with a Kuka LBR IIWA 14, which comes with its own safety challenges.  
\subsection{Quasi-Static Pick-and-Place Environment}

We evaluate the effectiveness of our proposed safety layer with visual constraints (\Secref{sec:visual_constraints}) on pick-and-place tasks on a Franka \textsc{fr3} platform. In these tasks, the \ac{rfm} is prompted to \textit{"grab the [object] and put it into the box"} where \textit{object} is taken from the set \{\texttt{apple, strawberry, banana, pear, tennis ball, baseball}\}.
As such, the goal is to safely grab the desired object from the table surface and place it into the box.
We evaluate three versions of this task with varying task complexity: \texttt{pick-fruit-easy}, \texttt{pick-ball-medium}, \texttt{pick-object-hard} (see \Cref{sec:task_descriptions} for more task details.
This common pick-and-place task represents various potential safety hazards, e.g., the robot can crash into the table while picking up the object, it can collide with obstacles in the scene, or attempt to leave the workspace.

We fine-tune \piZero{}~\citep{pi_0} on expert demonstrations of our above pick-and-place tasks to adapt it to the new environment.
We collect $300$ expert demonstrations via teleoperation following the data collection protocol from \citep{droid_2024}.
During data collection and evaluation, we randomize the position and orientation of the fruits and the box.
Further, we add and remove various \textit{distractor objects} within the scene (e.g., different boxes, cans, and bottles).
We outline further details on the setup in Appendix~\ref{sec:app_pick_and_place}. 

\begin{figure}[t]
    \centering
    \includegraphics[width=\textwidth]{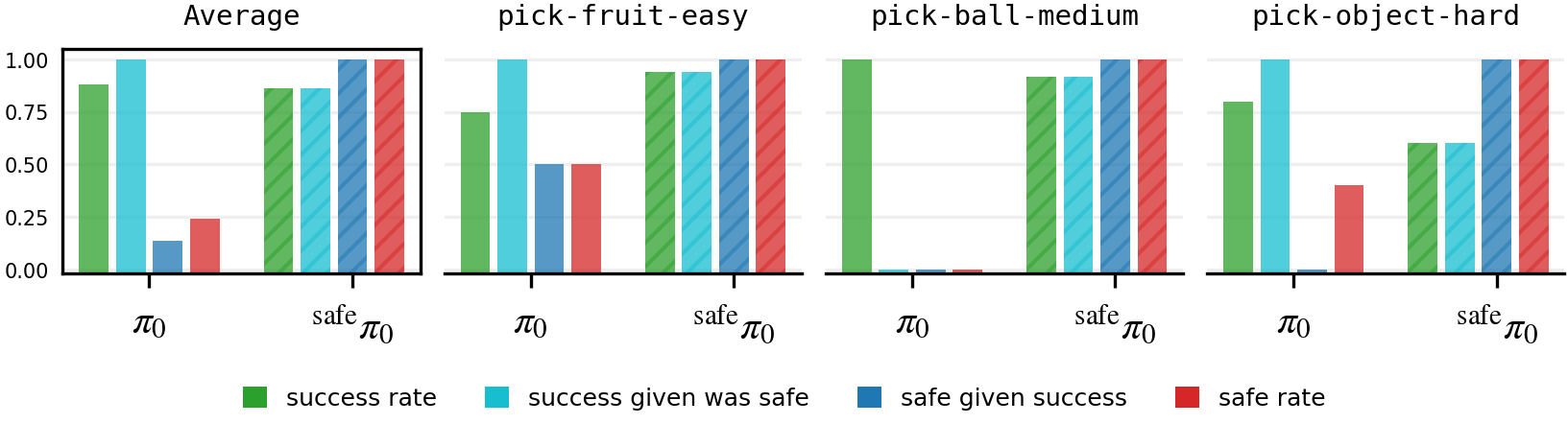}
    \caption{Results in the manipulation tasks. Dashed histograms indicate the \ac{rfm} combined with the \ac{atacom} safety layers, while the solid ones represent the vanilla \piZero{} model. 
    We report success rate, success rate for safe trajectories, percentage of safe trajectories among the successful ones, and normalized execution time. Results show that the safety layer does not impact heavily the success rate, while ensuring safety.}
    \label{fig:manipulation_results}
\end{figure}

The results for all of these experiments are presented in \Figref{fig:manipulation_results}. First, we evaluate each task's success rate, disregarding any safety considerations. 
Here, the success rate is evaluated visually, given the simplicity of the considered tasks. In general, our safety layer does not affect the performance heavily in terms of success rate. On average, the success rate is similar to that of the vanilla policy. Curiously, the safety layer increases the success rate in the pick-fruit-easy task. This counterintuitive result is because the safety constraint prevents collision with the table, allowing for a better grasp alignment. In two other tasks, the success rate is slightly lower, and it is particularly evident in the pick-object-hard task. This is reasonable, as imposing complex safety constraints makes the task execution harder.

To prove the benefits in terms of safety, we evaluate the task in terms of the success rate of safe trajectories (i.e., the ratio of safe trajectories that complete the task) and the safety rate of successful trajectories (i.e., the ratio of successful trajectories that are also safe). Results clearly show that our approach always ensures safe trajectories, while vanilla \piZero{} presents many constraint violations: in the pick-fruit-easy task, the model frequently collides with the table, whereas in the other two tasks, although the objectives are achieved, the robot often knocks over obstacles during execution.

\begin{wraptable}{r}{5.5cm}
    \vspace{-1.6em}
    \centering
    \begin{tabular}{@{}lll@{}}
    \toprule
                     & $\pi_0$ & $^\text{safe}\pi_0$ \\ \midrule
    pick-fruit-easy  & 26s      & 24s  \\
    pick-ball-medium & 23s      & 30s   \\
    pick-object-hard & 22s      & 22s   \\ \midrule
    Average          & 23s      & 25s  \\
    \bottomrule
    \end{tabular}
    \caption{Average time to task completion.
    Safe execution sometimes prolongs task duration.}
    \label{tab:success_time}
    \vspace{-1.1em}
\end{wraptable}

We also evaluate the performance in terms of execution time. The results are reported in Table~\ref{tab:success_time}. Specifically, we measure the execution time of successful trajectories. As expected, on average, the trajectories with the safety layer are slightly longer, particularly in the pick-ball-medium task. Again, we observe some performance gain in the pick-fruit-easy task, due to the lack of collisions with the table, which may cause the robot's end effector to drift, making the task harder for the unsafe policy.

\subsection{Dynamic Air Hockey Environment}
We empirically evaluate the proposed approach on a robot air hockey task. The objective is to hit a puck into the goal while adhering to multiple safety constraints, such as keeping the end-effector on the table surface, preventing the arm from colliding with the table, and ensuring joint position limits.  
We refer to \cite{liu2024_atacom_journal} for a detailed description of the experimental setup.
The policy's observation consists of language instructions, a goal image of the scene, and proprioceptive data in the form of joint positions, joint velocities, puck position, and puck velocity.
While not needed for safety, we fine-tune a pre-trained \ac{octo} \cite{octo_2024} policy using behavior cloning in a simulated \ac{mujoco}~\cite{todorov2012mujoco} environment. Importantly, we obtain the fine-tuning data by an expert policy that does not leverage \ac{atacom}.      
The policy outputs desired end-effector velocities in the x-y plane of the table surface, which are converted to joint velocities using inverse kinematics. The \ac{atacom} layer then maps these joint velocities to safe ones before passing them to a joint-space controller. We compare our safety-aware approach to an unsafe baseline, where the joint-space controller directly executes the unfiltered joint velocities. We evaluate the safety module for various fine-tuning checkpoints of \ac{octo} on the simulated system. Several deployment videos of \ac{octo} playing air hockey can be found on our project page.

\begin{figure}[t]
    \centering
    \begin{subfigure}[t]{0.48\textwidth}
        \includegraphics[width=\linewidth]{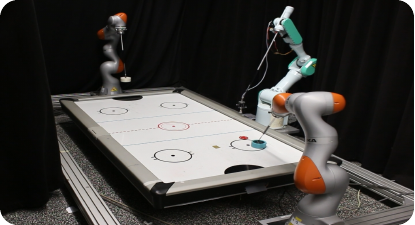}
    \end{subfigure}
    \begin{subfigure}[t]{0.49\textwidth}
        \includegraphics[width=\linewidth]{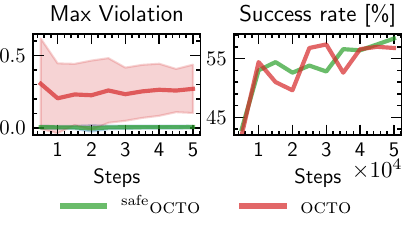}
    \end{subfigure}
    \caption{Safety violations of the \ac{octo} policy w/o the safety module on the air hockey hitting task for different checkpoints during the training phase. We report the maximum constraint violation and the success rate of the robot hitting the puck into the goal over 500 episodes in simulation. When the \ac{atacom} safety module is added, the policy remains compliant with safety constraints throughout fine-tuning, whereas the unmodified \ac{octo} policy continues to breach safety limits. Both policies progressively improve their success rates over the number of fine-tuning steps.}
    \label{fig:results}
\end{figure}

Our experimental results, presented in Figure~\ref{fig:results}, show that the \ac{octo} agent with the added safety module does not violate the safety constraints at deployment time. On the contrary, \ac{octo} without the added safety layer heavily violates the constraints, even though the fine-tuning data contains safe expert demonstrations.
Looking at the success rate, it is clear that using the safety module does not affect performance, as both approaches show similar behavior during the training phase. Importantly, while the fine-tuning data is not obtained with \ac{atacom}, we still obtain high success rates, suggesting that \ac{atacom} does not generate overly conservative control actions.

Finally, we deploy the safe policy in the real-world system. While the real-world deployment is affected by the sim-to-real gap, we still achieve reasonable performance while strictly enforcing the safety constraints. This shows that the approach is viable even in dynamic tasks and using nominal constraints, demonstrating the robustness of our approach against modeling errors.


\section{Related Work}
\label{sec:relatedwork}
This work considers ensuring safety during the deployment of robotic foundation models, which has been identified as an important open problem in the development of these policies \citep{firoozi2023foundation}.
As a robotic foundation model is abstractly like any other robot policy, albeit more capable, many existing safe control techniques can be directly applied.
In this work, we use \ac{atacom}~\citep{liu2022_atacom, liu2024_atacom_journal}, which augments the policy architecture to ensure safety in the action space.
\citet{zhang2025safevla} finetune a foundation policy using safe reinforcement learning techniques, which augment the objective with the constraints and Lagrangian multipliers.
Since safety is incorporated as an auxiliary objective, there is no guarantee that a constraint is obeyed at runtime and instead the policy pays a penalty for constraint violation.
Moreover, since a sample-inefficient on-policy RL method is used, the policy is trained and evaluated in simulation rather than the real world.

For complex multimodal policies, safety can also be interpreted as the performance of the policy to natural nonstationary perturbations of the real world during deployment.
\citet{majumdar2025predictive} approach safety from this robustness direction, and use generative models to achieve adversarial data augmentation to make the policy robust to sensing environmental factors like lighting; however, it does not adapt its behavior to dynamic environments.
\citet{hancock24byovla} use gradient information between the action and observations to augment the training data to alleviate the policy's sensitivity to the input space, as a proxy for robustness.
This line of research also extends to language conditioning, and ensuring safety via robustness to jailbreaking the language model component \cite{ravichandran2025safety}.

A separate line of work uses foundation models to ensure safety, using their real-world grounding to incorporate notions of `semantic' safety.
\citet{brunke2025semantically} use an \ac{llm} from semantic task descriptions to generate state and action constraints that are then enforced using control barrier function and safety filters. 
This approach is validated for actions generated from teleoperation and also diffusion policies.
\citet{ling2025impact} use a \ac{vla} to generate a cost map for motion planning, where the cost map automatically incorporates acceptable tolerances for practical obstacle avoidance. 
This is relevant for navigating complex cluttered scenes where certain objects, e.g., brittle and fragile, require greater care over objects that are soft or rugged.


\section{Discussion}
We propose a safety module that can be added as the final layer of an \acf{rfm} by leveraging domain-specific knowledge of the safe operation constraints.
We have shown that this layer can be added to a variety of existing \acf{rfm} architectures across a range of tasks, and ensure consistently safe execution with only small potential impacts to task performance.
We demonstrate the effectiveness of the safety layer by evaluating a \acf{vla} policy with \ac{bc} on an air hockey hitting task for which it is critical not to crash with the tabletop, and object manipulation tasks with obstacles.
We believe this architecture extension is necessary and crucial in the application of \acp{rfm} for everyday tasks and real-world deployment. 

Leveraging domain knowledge may seem counterintuitive for \acp{rfm}, as they show that rich behaviors can be obtained purely through large-scale datasets rather than handcrafted policies.
We believe that safety is inherently contextual information for a given task and robot configuration.
However, data-driven approaches leveraging this contextual cue are undesirable, as they would require collecting unsafe demonstrations in order for the policy to learn the difference between unsafe and safe behaviour. 
Therefore, we believe inductive biases are a more practical means of incorporating this contextual information, using minimal computational adjustments such as our proposed safety layer. 

While we emphasize that ensuring safety requires domain expertise, it can also be a demanding task to manually formulate all necessary safety constraints for a given task.
One intuitive research direction is to automate the process by leveraging the inherent knowledge of \acp{vlm}.
However, so far, \acp{vlm} have only been used to integrate semantic safety constraints such as `keep the cup uprigh' into an already existing set of constraints~\cite{updatingrobotsafetyrepresentations2024, brunke2025semantically}.
Beyond the formulation of safety constraints, it remains an open research question of how a more generalizable concept of safety can be formulated and applied across different embodiments, environments, and tasks.
One avenue, based on our safety layer, is to extend the `code-as-policies' paradigm \citep{liang2023code} to a `code-as-safety' approach, using \acp{llm} to automate the design of future safety layers.


\clearpage


\section*{Limitations}
The key limitation of our approach is that we assume the safety constraints as known. This may limit the applicability of the approach to open-world tasks, where it is not feasible to specify all the safety constraints a priori. However, our basic pipeline could be in principle extended with a simple object detection pipeline. Furthermore, it could be in principle possible to exploit the capabilities of \acp{llm} to generate a set of reasonable set of safety constraints for a given task.
This generation can be carried out offline, and therefore would not impact the execution time of the \ac{rfm} policy.
Another limitation of this paper is the limited evaluation of the policy to solve general tasks. While this is a very important topic, our focus is mostly on guaranteeing the safe execution of the policy. We believe that future models will generalize better across tasks and perform smoother and faster manipulation movements.
The last key limitation is that the setup requires multiple depth cameras to work properly and generate appropriate bounding boxes directly from perception. While this limits the applicability to general scenarios, many computer vision pipelines can generate robust bounding boxes, exploiting a moving camera. Furthermore, it may be necessary only to generate accurate bounding boxes from the side of the object perceived by the robot. 
For simplicity of the experimental setup, we left complex perception problems to future work.


\acknowledgments{}


\bibliography{bibliography}  

\newpage

\appendix

\section{Experimental Details: Pick-and-Place Tasks with Franka Robot}
\label{sec:app_pick_and_place}
We perform our vision-based pick-and-place experiments using a Franka Research 3 with a RH-P12-RN Robotis gripper, three Zed X mini cameras for scene capture, and a Meta Quest 3 with a single remote controller for data collection. We follow the hardware setup of DROID~\cite{droid_2024} and capture the scene with two external cameras to the left and right of the Franka robot and one gripper camera. The items to pick are selected from the classic YCB Benchmarks – Object and Model Set~\cite{ycb_object_set_2015}. 

\subsection{Expert Data Collection \& Fine-Tuning}
We adapt the open-source DROID pipeline~\cite{droid_2024}. Amongst others, we replace 
Polymetis~\cite{polymetis} with Franky~\cite{franky} to control our Franka robot. We collect over 300 trajectories for our experiments using the Meta Quest for teleoperation and showcasing pick-and-place behavior with different items in cluttered scenes. Importantly, the data is collected without \ac{atacom}~\cite{liu2022_atacom, liu2024_atacom_journal}, but showcases safe behavior. We convert the collected data into a LeRobot~\cite{lerobot} dataset and perform full finetuning of the \piZero{} base model~\cite{pi_0} on a single A100 GPU for 10.000 training steps with a batch size of 32. 

While we perform our experiments in joint space control, we collect data in Cartesian velocity space using the Meta Quest remote controller. We extract the necessary joint velocity information from the robot state to train the \piZero{} base model~\cite{pi_0}. Although joint velocity state and desired action values are expected to differ slightly, in practice, our finetuned joint velocity \piZero{} model performed well regardless.

\subsection{Pick-and-Place Franka Safety Layer}
We make the fine-tuned \piZero{} model safe by providing its joint velocity action output to our safety layer. Our safety layer is defined by several inequality constraints that define the constraint manifold. \ac{atacom} then generates safe actions by staying on the constraint manifold. For the workspace and bounding box constraints, the robot's hull gets approximated by several spheres $s_i$ as displayed in \Figref{fig:robot_spheres}. Each sphere is parameterized by its center position and its radius $({}^{\text{base}}\vx, r)$. The workspace constraints enforce that each sphere remains within the prescribed lower bounds ${}^{\text{base}}\vx_{\text{min}}$ and upper bounds ${}^{\text{base}}\vx_{\text{max}}$ of the workspace. In particular, for each sphere $s_i$ we require
\begin{align}
    \text{(Workspace)} && g_{s_i}(\vx) &= {}^{\text{base}}\vx_{\text{min}} - {}^{\text{base}}\vx + r_i  \leq 0 \\
                       && g_{s_i}(\vx) &= {}^{\text{base}}\vx - {}^{\text{base}}\vx_{\text{max}} + r_i \leq 0.
\end{align}
For the bounding box constraints, we calculate the distance between each sphere $s_i$ and the closest point on the bounding boxes surface 
\begin{align}
    \text{(Bounding Box)} && g_{s_i}(\vx) &= - || \alpha{}^{\mathrm{bb}}\vx - \vp\| \leq 0 \\
        && \quad \vp_j &= \mathrm{clip}( \alpha {}^{\mathrm{bb}}\vx, -{}^{\mathrm{bb}}\vh_j / 2, {}^{\mathrm{bb}}\vh_j / 2) \notag \\
        && \quad\alpha &= 1 - r/ \| {}^{\mathrm{bb}}\vx\| \notag,
\end{align}
where ${}^{\text{bb}}\vx$ is the sphere's center position in the bounding box (bb) frame and ${}^{\text{bb}}\vh$ corresponds to the extent of the bounding box. A third constraint category keeps the joints $\vq$ within their limits $\vq_{\mathrm{min}}, \vq_{\mathrm{max}}$ 
\begin{align}
    \text{(Joint Limits)} && g(\vx) &= ((\vq - \bar\vq)^2 / \Delta q^2) - 1 \leq 0 \\
    && \bar\vq &= (\vq_{\mathrm{max}} + \vq_{\mathrm{min}}) / 2 \notag  \\    
    && \Delta\vq &= (\vq_{\mathrm{max}} - \vq_{\mathrm{min}}) / 2.  \notag
\end{align}

\subsection{Evaluation Task Descriptions}
\label{sec:task_descriptions}
We evaluate our Franka pick-and-place safety layer on three evaluation tasks which we further describe in the following paragraphs. Selected hyperparameters of $\pi_{0}$ and \ac{atacom} can be found in Table \ref{tab:pick_and_place_task}.

\begin{wraptable}{r}{4.5cm}
    \centering
    \caption{Parameter Selection for \\ 
    the Pick-and-Place Tasks}
    \label{tab:pick_and_place_task}
    \begin{tabular}{l r}
    \toprule
    Hyperparameter & Value \\
    \midrule
    $\pi_0$ & \\
    \midrule
    Model & Base \\
    Control frequency & 15hz \\
    Action chunk size & 32 \\
    \midrule
    \ac{atacom} & \\
    \midrule
    Control frequency & 60 hz \\
    Slack function     &  exp\\
    Slack beta     &  10\\
    Slack tolerance & 0.001 \\
    Drift clipping & True \\
    \bottomrule
    \end{tabular}
    \vspace{-5em}
\end{wraptable}

\paragraph{pick-fruit-easy.}
This task requires picking a specified (plastic) fruit off the table and placing it in the box.
We consider the task to be successful when the robot manages to place the specified fruit into the box within the maximum episode length of $1000$ timesteps.
The potential danger is to collide with the table surface, especially for smaller fruits like the strawberry. 

\paragraph{pick-ball-medium.}
This task requires picking up a tennis ball off the table and placing it within the box. The ball is placed close ($\sim10$cm) from a large cardboard box, which serves as an obstacle that the robot needs to avoid during the pickup.
The robot needs to avoid collisions with the obstacle as well as the table simultaneously.

\paragraph{pick-object-hard.}
This task requires the robot to pick up a specified object off the table and place it in the box. We add up to $3$ obstacles into the workspace that the robot needs to avoid during operation.

\section{Experimental Details: Pushing Task on Air Hockey Table}
We perform our AirHockey experiments in the \ac{mujoco}~\cite{todorov2012mujoco} simulator with a fine-tuned \ac{octo}~\cite{octo_2024} model. The input to the model consists of the tuple $\vx = (\vo, \vq, \dot\vq, \vp, \dot\vp, \vg_{\mathrm{img}})$, which comprises a third person visual observation $\vo$, joint positions $\vq\in\sR^{7}$ and the joint velocities $\dot\vq\in\sR^{7}$, the puck's horizontal and rotational position $\vp = [x, y, \theta]^\intercal$ as well as velocity $\dot\vp$, and a task to achieve, given as goal image $\vg_{\mathrm{img}}$. We fine-tune \ac{octo}~\cite{octo_2024} to predict desired Cartesian velocity action chunks in the 2D Cartesian space of the mallet end-effector on the AirHockey table. The model gets evaluated on the task of hitting a randomly initialized puck into the goal on the other side of the AirHockey table.

\subsection{Expert Data Collection and Fine-Tuning}
We collect our fine-tuning data with an expert policy, which does not leverage ATACOM~\cite{liu2022_atacom, liu2024_atacom_journal}. Data is collected independently of our safety layer, as we do not need fine-tuning data to become safe, but to increase the performance of the policy in the new environment. We spawn the puck randomly in a rectangle in front of the robot with a zero initial velocity. The expert policy generates trajectories using a classical planning approach to hit the puck into the goal on the other side of the table. In total, we collect 500 trajectories for fine-tuning. 
We fine-tune \ac{octo} on the collected data for 50.000 gradient steps using a batch size of 256. As we provide additional puck state information, we do not need to deal with partial observability and choose an observation history of zero. Initial evaluations have shown that we obtain the best puck-hitting results with an action chunk size of 16.

\subsection{AirHockey Safety Layer}
We guarantee safe deployment of the fine-tuned \ac{octo}~\cite{octo_2024} model by feeding predicted 2D Cartesian actions into our ATACOM safety layer. As our safety layer works in joint space, we first convert the Cartesian velocities into joint velocities using inverse kinematics. ATACOM morphs these joint velocities into safe ones by acting on the tangent space of the constraint manifold. Several inequality constraints define the constraint manifold. Our table surface constraints 
\begin{align}
    \text{(Table surface)} && g_1(\vx) &= -z_{\text{ee}} + z_{\text{low}} \leq 0 \\
    && g_2(\vx) &= z_{\text{ee}} - z_{\text{high}} \leq 0
\end{align}
ensure that the attached mallet stays within the table height bounds $z_{\text{low}}, z_{\text{high}}$. With the additional link constraints
\begin{align}
    \text{(Link constraints)} && g_3(\vx) &= -x_{\text{ee}} + x_{\text{low}} \leq 0 \\
                              && g_4(\vx) &= -y_{\text{ee}} + y_{\text{low}} \leq 0 \\
                              && g_5(\vx) &= y_{\text{ee}} - y_{\text{high}} \leq 0 \\
                              && g_6(\vx) &= -z_{\text{wrist}} + z_{\text{wrist}_{\text{low}}} \leq 0 \\
                              && g_7(\vx) &= -z_{\text{elbow}} + z_{\text{elbow}_{\text{low}}} \leq 0,
\end{align}
we keep the mallet within the table bounds $x_{\text{low}}, y_{\text{low}}, y_{\text{high}}$ as well as the wrist and elbow above lower height bounds $z_{\text{wrist}_{\text{low}}}, z_{\text{elbow}_{\text{low}}}$.
The last constraint category keeps the joints $\vq$ within their limits $\vq_{\mathrm{min}}, \vq_{\mathrm{max}}$
\begin{align}
    \text{(Joint limits)} && g_{8...14}(\vx) &= ((\vq - \bar\vq)^2 / \Delta q^2) - 1 \leq 0 \\
    && \bar\vq &= (\vq_{\mathrm{max}} + \vq_{\mathrm{min}}) / 2 \notag  \\    
    && \Delta\vq &= (\vq_{\mathrm{max}} - \vq_{\mathrm{min}}) / 2.  \notag
\end{align}

\subsection{Evaluation \& Real-World Deployment}
We evaluate every \ac{octo}~\cite{octo_2024} fine-tuning checkpoint for 500 episodes in our \ac{mujoco}~\cite{todorov2012mujoco} environment. Important hyperparameters of our evaluation can be found in Table \ref{tab:pushing_task}. 

After evaluation in simulation, we deployed ${}^{\text{safe}}\ac{octo}$ on our real-world AirHockey setup. \ac{octo}, without our safety layer, was not secure enough to be deployed in real world. Amongst others is the risk that \ac{octo} strongly hits the AirHockey table while performing the task. As the policy was trained in simulation with additional puck state information, we use the OptiTrack system to track the puck's position and velocity. Using proprioception and puck state information, we reconstruct the real-world state in \ac{mujoco}~\cite{todorov2012mujoco} to obtain the simulated visual input observation. Videos of ${}^{\text{safe}}\ac{octo}$ playing AirHockey in the real world can be found on our project page. 

\begin{table}[h]
    \centering
    \begin{tabular}{l r}
    \toprule
    Hyperparameter & Value \\
    \midrule
    \ac{octo} & \\
    \midrule
    Model & octo-small-1.5 \\
    Control frequency & 12.5 hz \\
    Action chunk size & 16 \\
    \midrule
    \ac{atacom} & \\
    \midrule
    Control frequency & 50 hz \\
    Slack function     &  exp\\
    Slack beta     &  2  \\
    Slack tolerance & $1e^{-6}$ \\
    Drift clipping & True \\
    \bottomrule
    \end{tabular}
    \vspace{0.5em}
    \caption{Parameter Selection for 
    the AirHockey Pushing Task}
    \label{tab:pushing_task}
\end{table}

\end{document}